\documentclass[10pt]{IEEEtran}
\usepackage[utf8]{inputenc}
\usepackage{amsfonts}
\usepackage{amssymb}
\usepackage{amsmath}
\usepackage{microtype}
\usepackage{graphicx}
\usepackage{booktabs}
\usepackage{hyperref}
\usepackage{balance}
\usepackage{multirow}
\usepackage{footnote}
\usepackage{tablefootnote}
\usepackage{gensymb} 
\usepackage{commath} 
\usepackage{bm} 
\usepackage{siunitx} 
\usepackage{xcolor} 
\usepackage[noadjust]{cite} 
\usepackage{threeparttable} 
\usepackage{algorithmic}
\usepackage{algorithm}
\usepackage[algo2e]{algorithm2e} 
\usepackage{caption}
\makesavenoteenv{table}
\makesavenoteenv{table*}
\makesavenoteenv{tabular}

\newcommand{\R}[1] {\textnormal{#1}}

\begin{document}

\title{Deformable Radar Polygon: A Lightweight and Predictable Occupancy Representation for Short-range Collision Avoidance}

\author{Xiangyu Gao, Sihao Ding, Harshavardhan Reddy Dasari
\thanks{X. Gao is with the Department of Electrical and Computer Engineering, University of Washington, Seattle, WA, 98195. (Email: {xygao@uw.edu})} 
\thanks{S. Ding, and H. Dasari are with the Volvo Cars, Sunnyvale, CA, 94085. (Email: {\{sihao.ding, harshavardhan.reddy.dasari\}@volvocars.com})}
\thanks{Corresponding Author: Xiangyu Gao}
}

\maketitle

\begin{abstract}
Inferring the drivable area in a scene is crucial for ensuring a vehicle avoids obstacles and facilitates safe autonomous driving. In this paper, we concentrate on detecting the instantaneous free space surrounding the ego vehicle, targeting short-range automotive applications. We introduce a novel polygon-based occupancy representation, where the interior signifies free space, and the exterior represents undrivable areas for the ego-vehicle. The radar polygon consists of vertices selected from point cloud measurements provided by radars, with each vertex incorporating Doppler velocity information from automotive radars. This information indicates the movement of the vertex along the radial direction. This characteristic allows for the prediction of the shape of future radar polygons, leading to its designation as a ``\textit{deformable radar polygon}". We propose two approaches to leverage noisy radar measurements for producing accurate and smooth radar polygons. The first approach is a basic radar polygon formation algorithm, which independently selects polygon vertices for each frame, using SNR-based evidence for vertex fitness verification. The second approach is the radar polygon update algorithm, which employs a probabilistic and tracking-based mechanism to update the radar polygon over time, further enhancing accuracy and smoothness. To accommodate the unique radar polygon format, we also designed a collision detection method for short-range applications. Through extensive experiments and analysis on both a self-collected dataset and the open-source RadarScenes dataset, we demonstrate that our radar polygon algorithms achieve significantly higher IoU-gt and IoU-smooth values compared to other occupancy detection baselines, highlighting their accuracy and smoothness.
\end{abstract}


\section{Introduction}
The detection of occupancy is a crucial aspect of comprehending road scenes in the context of autonomous driving. It encompasses information pertaining to the drivable area and road obstacles, playing a vital role in ensuring safe autonomous navigation. An ideal occupancy representation assigns a pixel value of 1 to locations occupied by obstacles and a pixel value of 0 to locations representing free space. In numerous short-range applications (i.e., covers less than \SI{30}{m} range \cite{srr}), such as vehicle back-off planning in parking lots, the focus is on the instantaneous free space surrounding the sensor \cite{gao2019experiments, gao2021mimosar}. To address this need, our objective is to develop a precise and lightweight occupancy representation specifically tailored for the detection of instantaneous free space in short-range applications.

Free space detection can be achieved through various sensor modalities, including ultrasound and cameras. While cameras excel in image-based free space detection, they are limited by challenging weather and lighting conditions (such as night, rain, and snow) \cite{gao2019experiments}, and they may face accuracy limitations in depth estimation using a single device. On the other hand, ultrasound sensors have been utilized for short-range occupancy guidance, but their performance can be unstable, leading to high false alarm rates due to variations in air properties like temperature and wind \cite{ti_uss}. Consequently, we opt for mmWave radars \cite{ramp} for free space detection. In recent years, radars have gained prominence in autonomous driving due to their affordability, range-Doppler-angle localization capabilities \cite{gao2019experiments, gao2021mimosar}, and resilience under adverse weather conditions \cite{gao2021perception}.

Various approaches have been proposed to address occupancy and free space detection. Traditional methods involve occupancy grid mapping, employing Bayesian filtering and hand-crafted inverse sensor model (ISM) functions \cite{8443484, derive_spat_occup, 7535495, 9299052, 8916897, 9022091, 7117922, 9294626, Li2018HighRR}. This approach accumulates confidence in occupancy over time for a grid map area, offering accurate estimations for static objects with known ego-vehicle motion values. However, occupancy grid techniques have limitations in highly dynamic environments. These techniques often assume a static environment, making it challenging to handle moving objects. Grid cells represent static obstacles, requiring frequent classification and updates to reflect changes, which can be computationally expensive. Without timely updates, moving objects can cause false alarms and missed regions on the grid map. Besides, occupancy grid techniques require substantial memory consumption during formation and updating over time, making it less cost-effective for short-range applications \cite{5548091, 8813839}. Beyond traditional methods, recent works treat occupancy detection as a semantic segmentation problem, employing deep learning (DL) models \cite{gao2022learning} to predict the occupancy status for each pixel or cell in a map. DL-based methods, often using large neural networks, surpass traditional methods in performance but require substantial labeled training sets and intensive training and computing processes \cite{Sless_2019_ICCV, xie2022deepvs, 10273590, 9341308, xie2021vitalhub, 8793263, xie2022passive}. Additionally, concerns persist regarding the generation capabilities of DL-based methods due to limitations in training and testing sets.

\begin{figure}
\centering
\includegraphics[width=0.45\textwidth]{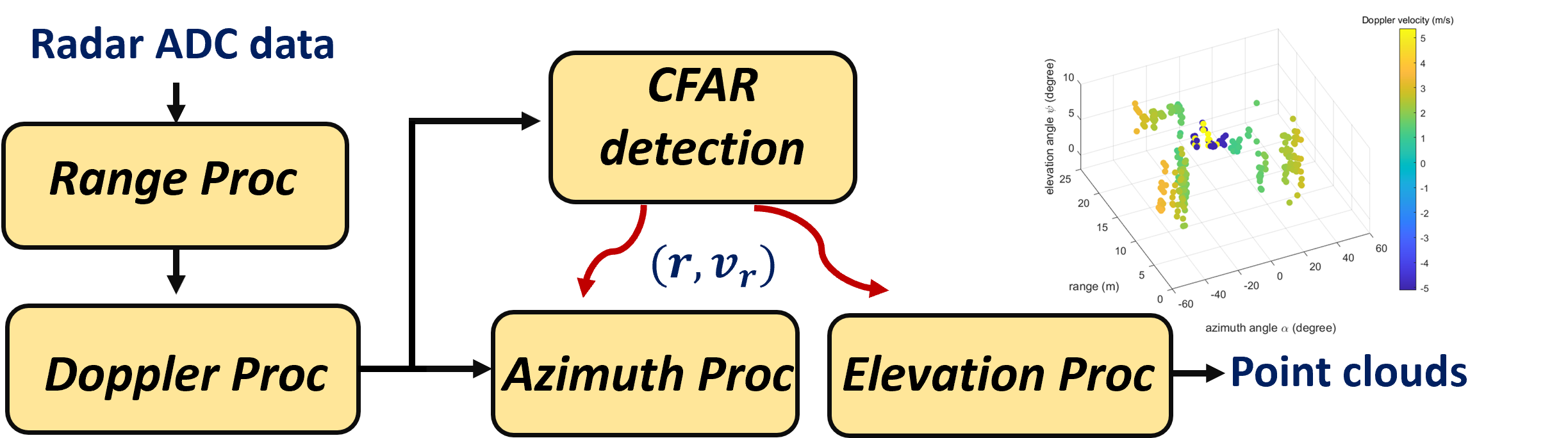}
\caption{A high-level processing diagram for obtaining the radar point clouds from raw ADC data \cite{gao2023static}.}
  \label{fig:point}
\end{figure}

This paper introduces a novel occupancy representation tailored for short-range applications leveraging automotive radars, termed the \textit{``deformable radar polygon"}. The radar polygon format defines a polygon-shaped region where the interior signifies free space, and the exterior represents undrivable areas for the ego-vehicle. This format prioritizes accurate modeling of the free space near the ego vehicle, aligning with the requirements of short-range applications that emphasize proximity to surrounding obstacles. With point clouds extracted from raw radar data (an example pipeline is shown in Fig.~\ref{fig:point}), the radar polygon comprises vertices selected from these point cloud measurements. An intriguing aspect of the radar polygon is that each vertex includes Doppler velocity information from automotive radars, indicating the movement of the vertex (shrinkage or expansion) along the radial direction. This characteristic enables the prediction of the shape of future radar polygons, offering valuable insights for downstream applications such as route planning and collision avoidance. Due to these predictive capabilities, we refer to the radar polygon as \textit{``deformable"}. Beyond its predictive property, the radar polygon is computationally and storage-efficient, requiring the storage of a limited number of vertices and their confidence levels, in contrast to processing a large number of cells across the entire map (e.g., occupancy grid).

To construct the radar polygon, we present two approaches: the first is the \textit{``basic radar polygon"} algorithm, which selects vertices independently for each time slot using heuristic methods, with signal-to-noise ratio (SNR)-based evidence employed for vertex fitness verification. The second approach is the \textit{``radar polygon update"} algorithm, an advancement of the basic radar polygon method that takes into account information from previous frames. To enhance the system's performance in terms of accuracy and smoothness, we introduce a probabilistic and tracking-based mechanism for updating the radar polygon over time. Evaluation results demonstrate that the proposed radar polygon methods exhibit low memory consumption, fast processing, and high smoothness, and achieve a comprehensive and accurate representation of free space. Additionally, we introduce a collision detection algorithm to identify potential collisions outside the polygon region, leveraging the odd-even principle \cite{10.1145/368637.368653}. 

In summary, the key contributions are four-folded:
\begin{itemize}
\item We proposed a polygon-based occupancy representation to model the free space around the ego vehicle for short-range applications. A basic radar polygon formation algorithm was proposed, which selects polygon vertices for each time slot, with SNR-based evidence employed for vertex fitness verification.
\item Beyond the basic radar polygon, we proposed the radar polygon update algorithm, which takes a probabilistic and tracking-based mechanism for updating the radar polygon over time to enhance the accuracy and smoothness.
\item We introduced the novel concept of the ``deformable radar polygon," which predicts the future shape of the free space polygon based on the Doppler velocity of its vertices. Additionally, we devised a method for collision detection on radar polygons by determining if points are situated inside the testing polygon. 
\item To assess the performance of our proposed algorithms, we conducted comprehensive evaluations and analyses on both our self-collected dataset and the publicly available RadarScenes dataset. The results show significantly improved IoU-gt and IoU-smooth values compared to baseline methods.
\end{itemize}

\section{Related Work}
The process of forming LiDAR-based occupancy grid maps \cite{9022091} typically involves utilizing a delta function for the ISM. This choice is influenced by the dense and accurate nature of LiDAR point data. Conversely, when creating a radar occupancy grid map, a more prevalent approach is to employ a Gaussian variant of the delta function for ISM. This adaptation accounts for the sparser and noisier characteristics of radar data. Notably, in works such as \cite{7117922, Li2018HighRR}, the Gaussian ISM has been further enhanced by incorporating detection probability calculations, considering the plausibility of range, angle, and amplitude for each radar measurement. In the realm of ISM improvement, \cite{8443484} introduces ego-motion velocity-dependent parameters to the algorithm. This enhancement dynamically adjusts uncertainty ellipses based on ego velocity, assigning higher values when velocity is high and detections are scarce. To address the inherent uncertainty in radar measurements, \cite{7535495} proposes converting SNR to the probability of detection using the Swerling-1 model within the ISM framework. Adapting to diverse application scenarios, existing radar occupancy grid map methods have been extended to facilitate numerically efficient computations in various dimensions \cite{8813839, 7535495}. Moving beyond traditional occupancy grid maps, \cite{8793503} introduces a novel approach using 3D Mesh to represent free space. This method balances system precision and efficiency, demonstrating its versatility in handling complex scenarios. Furthermore, \cite{9740418} focuses on predicting a drivable path for self-driving vehicles within the field of view (FOV) of a radar sensor using the dynamic Gaussian distributions for occupancy indication. In the realm of polygon-based representation, Meerpohl \textit{et al.} \cite{MEERPOHL2019368} employ raycasting to delineate the boundaries of free space within a grid map, resulting in a closed polyline. Notably, their approach relies on the prior formation of a grid map, a prerequisite before executing the algorithm. Ziegler \textit{et al.} \cite{6856581} adopt a strategy where occupied areas on the left and right bounds of the driving corridor are defined as sets of convex hulls, particularly useful for path planning. Additionally, Kuan \textit{et al.} \cite{kuan1985natural} present an algorithm that, given a set of polygonal obstacles in space, decomposes the free space into nonoverlapping geometric-shaped primitives suitable for path planning.

Current ISM-based algorithms heavily rely on radar detection and are predominantly hand-crafted \cite{10273590}. Inferring occupied space based on radar detections poses notable challenges due to data sparsity and environment-dependent noise, such as multipath reflections. Recent developments, moving away from hand-crafted models, involve the utilization of DL-based methods in occupancy grid mapping, leveraging a data-driven approach \cite{9299052, Sless_2019_ICCV, 10273590, 9341308, 8793263}. DL-based ISM models consider the spatial coherence of all radar detection points to predict the scene context \cite{9341308, 8793263}. The goal is to enhance system performance through an end-to-end model. Typically, deep ISM methods treat the occupancy status as a semantic segmentation problem, predicting the semantic class (i.e., occupied or free space) or the occupancy probability for each pixel in the occupancy grid.

Building upon the generated radar occupancy grid, various downstream applications have been explored. These applications include free space detection through image analysis and dynamic b-spline contour tracking \cite{6338636}, detection of parallel-parked and cross-parked vehicles \cite{6856568}, and identification of available parking spaces \cite{7918864}, among others. Beyond occupancy mapping, there is a growing interest in vehicle perception using radars within the research community. This interest is fueled by the widespread availability of radars in off-the-shelf cars, their relatively low cost, their capability for 3D/4D estimation \cite{gao2019experiments, gao2021mimosar}, and their robustness under challenging weather conditions \cite{gao2021perception}. Novel applications, such as super-resolution imaging \cite{gao2021mimosar}, vital sign monitoring \cite{xie2021signal, xie2021fusing}, and road object detection \cite{ramp}, have been proposed to leverage the full potential of automotive radars.


\section{Basic Radar Polygon Formation}
\subsection{Problem Formulation}
In a set of radar detections denoted as $\mathbf{E}$, each detection $e$ encompasses measurements in two dimensions $(x, y, z)$, Doppler velocity $v$, and SNR. Our goal is to select a subset of points, denoted as $\mathbf{E}^\prime$, from these detections to serve as vertices for constructing an occupancy polygon. The enclosed area of this polygon is considered free space. Our objective is to accurately determine the free space, minimize the number of selected points, and mitigate the interference from false alarms. To solve it, we propose the following heuristic solution.

\begin{figure}
\centering
\includegraphics[width=0.48\textwidth, trim=1 2 1 1,clip]{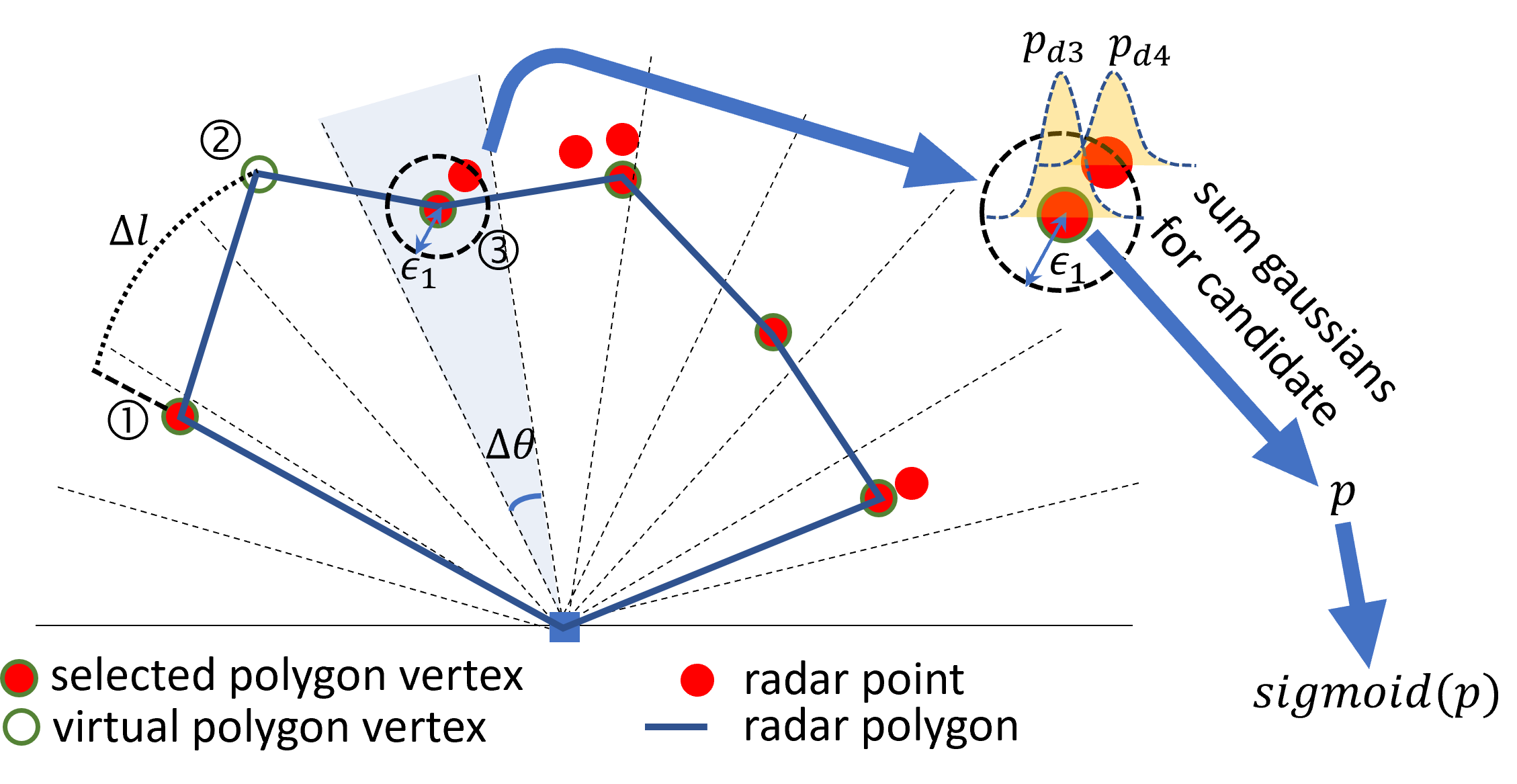}
\caption{Procedures of forming a basic radar polygon. After sampling the FOV with $\Delta \theta$, a vertex (point \textcircled{\raisebox{-0.9pt}{3}}) is selected based on normalized evidence probability, or a virtual vertex (point \textcircled{\raisebox{-0.9pt}{2}}) is created for each sampling sector.}
\label{fig:polygon_form}
\end{figure}

\begin{figure*}
\centering
\includegraphics[width=0.95\textwidth]{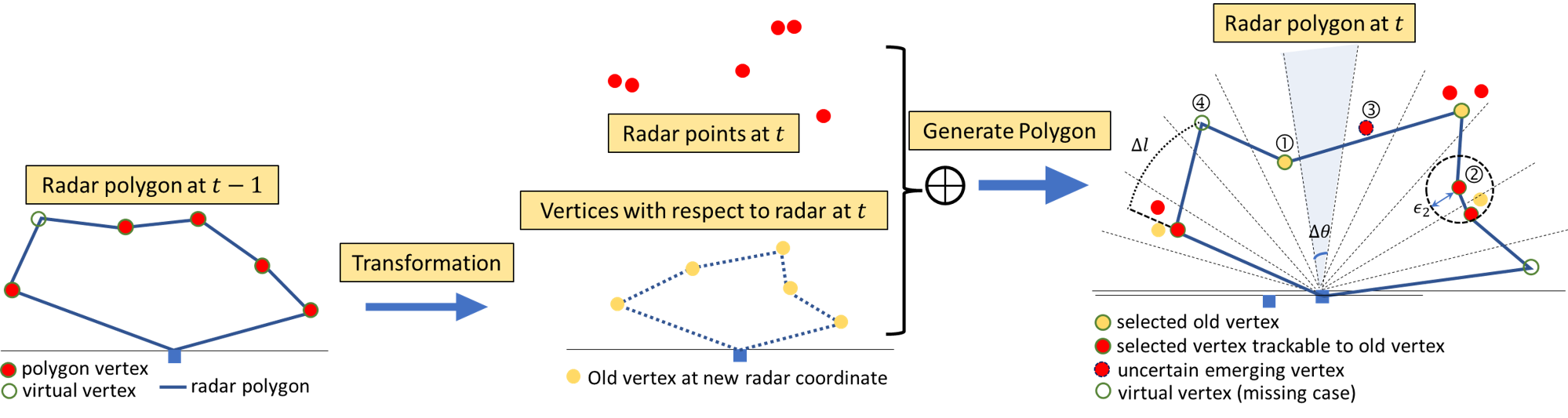}
\caption{The radar polygon update algorithm follows this processing diagram: First, the radar polygon from the previous frame \(t-1\) is transformed to the new coordinate system centered at the radar at \(t\), using the vehicle pose. The vertices in the transformed polygon are then integrated into the latest radar measurements for radar polygon generation. The formation steps follow the description in Section~\ref{sec:pol_ism}. Specifically, there are four types of vertices: selected old vertex (point \textcircled{\raisebox{-0.9pt}{1}}), selected vertex trackable to old vertex (point \textcircled{\raisebox{-0.9pt}{2}}), uncertain emerging vertex (point \textcircled{\raisebox{-0.9pt}{3}}), and virtual vertex (point \textcircled{\raisebox{-0.9pt}{4}}). The uncertain emerging vertex will not be used for polygon formation in the current frame but will be stored and associated with future frames multiple times before being activated.}
  \label{fig:poly_update}
\end{figure*}

\subsection{Basic Radar Polygon Formation \label{sec:pol_form}}
In this section, we present a method for constructing a basic radar polygon in each time slot, independent of information from previous slots. This approach utilizes SNR-based evidence verification to identify suitable polygon vertices. Before initiating this process, we clean the data before feeding the radar point clouds to the algorithm by filtering out the points with very large heights (i.e., larger than $z_\text{max}$) and very small heights (i.e., smaller than $z_\text{min}$). Points with extreme heights often have less amplitude and could be false alarms resulting from the energy leakage of strong-amplitude targets. Besides, these points are outside the vehicle's safe passing region, so disregarding them does not significantly affect the free space detection. In our case, we have chosen $z_\text{max}$ to be \SI{3}{m} and $z_\text{min}$ to be \SI{-1.5}{m}. Since our focus is on the occupancy status in a 2D bird's-eye-view (BEV), we project the processed points onto the 2D $x$-$y$ plane by discarding the $z$ dimension. 

A polygon comprises multiple vertices, and our approach involves identifying a suitable vertex for each azimuth direction, as depicted in Fig.~\ref{fig:polygon_form}. To achieve this, we evenly sample the entire FOV using a fixed angle interval $\Delta \theta$. A vertex is then determined from radar points within each specific sampling sector (e.g., the light blue area in Fig.~\ref{fig:polygon_form}). The criteria for selecting the vertex in each sampling sector are defined as follows.

Firstly, choose a \textit{candidate vertex} from all radar points within the sampling sector. The points are sorted based on their distance $d$ defined as $d=\sqrt{x^2+y^2}$. The point with the shortest distance is selected as a candidate, and its validity as a \textit{valid vertex} is then verified. If it doesn't meet the criteria, the process continues with the second-shortest-distance point, and so forth.

Verification involves examining the SNR values of neighboring points within a distance of $\epsilon_1$ from the candidate vertex. This step aims to prevent low-quality points, such as false alarms, from being selected as the polygon vertex. Each neighboring point is assigned a detection probability $p_d$ based on its SNR \cite{7535495} value, utilizing the Swerling-1 model:
\begin{equation}
\label{eq:det_prob}
p_{d}=p_{\R{fa}}^{\frac{1}{1+\text{SNR}}}
\end{equation}
\noindent where $p_{\R{fa}}$ represents the false alarm rate used in point cloud extraction. The occupancy evidence map is established based on the detection probability $p_d$ and the point location $\bm{\mu}=(x_0,y_0)$. It is defined by a Gaussian distribution, denoted as $p_d\cdot \mathcal{N}(\bm{\mu},,\bm\Sigma)$, with $\bm\Sigma$ as a fixed variance parameter \cite{derive_spat_occup}. To ensure that the evidence map adequately covers neighboring points, and considering that $99.7\%$ of the energy of a normal distribution lies within the $3$-sigma region, we set $\bm \Sigma= (\epsilon_1/3)^2 \bm I$, where $\epsilon_1$ is the specified distance. The evidence map assigns an evidence value to any location $(x,y)$ on the map according to the Gaussian distribution $p_d\cdot \mathcal{N}(\bm{\mu}, \bm\Sigma)$. Closeness to $\bm{\mu}$ results in a higher evidence value, approaching $p_d$.

When multiple points are present in the $\epsilon_1$-neighborhood of the candidate point, each point generates a Gaussian evidence map in the $x$-$y$ plane, resulting in multiple overlapping evidence maps. As depicted in Fig.~\ref{fig:polygon_form}, the cumulative Gaussian evidence from all maps at the candidate vertex's location is computed. The overall evidence, denoted as $p$, is obtained by summing these Gaussian evidences. Subsequently, $p$ is normalized within the range $[0,1]$ using the \textit{sigmoid} function:
\begin{equation}
\label{eq:sigmoid}
\resizebox{.83\hsize}{!}{
$\tilde{p}=\textit{sigmoid}(p)=\frac{1}{2} + \frac{1}{2}\left[1+\exp \left(\frac{-\left(p-\bar{p}\right)}{\sigma_{p}}\right) \right]^{-1}
$}
\end{equation}
\noindent where $\bar{p}$ and $\sigma_p$ represent the shift and scaling factors. The \textit{sigmoid} function is applied to normalize $p$ to $\tilde{p}$ within the range $[0,1]$. Only when the calculated $\tilde{p}$ exceeds the threshold $p_\R{thr}$ do we confirm the candidate as the vertex. Otherwise, we select the next-shortest-distance point as the candidate and repeat the verification above process. It's important to note that the use of the $\epsilon_1$ distance threshold helps limit the number of points considered in the verification process, thereby alleviating computational load.

Secondly, following candidate verification for all points in the sampling sector, in the absence of any satisfying points, a \textit{virtual vertex} is created. This virtual vertex is positioned at the intersection of the sector center and the boundary (e.g., point 2 in Fig.~\ref{fig:polygon_form}), characterized by zero Doppler velocity and zero SNR. When examining two adjacent non-virtual vertices (e.g., points 1 and 3), if the cross-range length $\Delta l$ (arc length) between them exceeds the predefined threshold $l_\R{thr}$, a virtual vertex is not permitted in the middle sampling sector between them. This restriction is implemented to prevent the formation of small spikes associated with the virtual vertex.

Finally, all valid vertices, along with the origin point, are sequentially connected to form the basic radar polygon.

\section{A Probabilistic and Tracking-based Algorithm for Radar Polygon Update \label{sec:pol_ism}}
While the basic radar polygon formation algorithm yields an occupancy polygon around the ego-vehicle for each timeslot, it lacks consideration of previous frames. To enhance confidence in the polygon boundary and address gaps in sensor coverage, we propose a probabilistic and tracking-based algorithm for updating the radar polygon over time. The underlying assumption is that vertices in each sampling sector can be tracked across timeslots when associated with the same target. Leveraging this, we employ the ISM to calculate a posterior probability or confidence for each vertex, indicating its validity. Unlike the traditional use of ISM in fixed-cell occupancy grid formation, our approach accommodates unfixed vertex locations, functioning effectively under the specified assumption. Mathematically, the problem is defined as follows: given the polygon vertices set $E^\prime_{t-1}$ and the confidence set $L_{t-1}$ for the vertices at time $t-1$, along with the noisy radar measurements $E_{t}$ and the vehicle pose $Z_t$ at time $t$, the objective is to output the new polygon vertices $E^\prime_{t}$ and their associated confidence $L_{t}$ for the current time $t$.

To address this, we initially translate the locations of previous vertices $E^\prime_{t-1}$ to the latest coordinates at time $t$ using the provided vehicle pose $Z_t$. Assuming $Z_t$ is represented as $(x_t, y_t, \phi_t)$ -- where $x$ and $y$ denote global position, and $\phi$ is the orientation -- this transformation compensates for the vehicle's motion $(x_t-x_{t-1}, y_t-y_{t-1})$ and orientation change $\phi_t-\phi_{t-1}$. Subsequently, we integrate the updated vertices $E^\prime_{t-1}$ into the latest radar measurements $E_t$ for further processing, i.e., $E_t= E_t \cup E^\prime_{t-1}$. The next step involves applying the first stage of the basic radar polygon formation algorithm (as detailed in Section~\ref{sec:pol_form}) to $E_t$ to select candidate vertices $e^\prime_t$ for each sampling sector. However, the processing of these candidate vertices occurs under four distinct cases:
\begin{itemize}
    \item \textit{An old vertex}: This occurs when the candidate vertex $e^\prime_t$ coincides with a point from the previous vertex set $E^\prime_{t-1}$. If the candidate's confidence $\ell_{t-1}$ is negative, indicating it is no longer valid, we discard it and seek a new candidate in the same sampling sector. For positive confidence, we add it to the set $E^\prime_t$ but impose a confidence penalty of $-l_{\text{pen}}$ to account for the possibility of indefinite persistence.
    
    \item \textit{Trackable to an old vertex}: This situation arises when the candidate $e^\prime_t$ is not part of $E^\prime_{t-1}$ but is close to the latest location of the previous vertex $e^\prime_{t-1}$ within a threshold $\epsilon_2$. In this case, we assume $e^\prime_t$ and $e^\prime_{t-1}$ belong to the same target and are trackable. The update of confidence or posterior probability $\ell_t$ for $e^\prime_t$ is based on Bayes' theorem \cite{Li2018HighRR} and the confidence $\ell_{t-1}(e_{t-1}^\prime)$ of the previous vertex $e^\prime_{t-1}$:
    \begin{equation}
    \label{eq:bayes_log}
    \ell_{t}(e_t^\prime)=\ell_{t-1}(e_{t-1}^\prime)+\log \frac{\tilde{p}\left(e_t^\prime \mid E_{t}, Z_{t}\right)}{1-\tilde{p}\left(e_t^\prime \mid E_{t}, Z_{t}\right)}-\ell_{o}
    \end{equation}
    where $\tilde{p}\left(e_t^\prime \mid E_{t}, Z_{t}\right)$ is the normalized occupancy evidence of $e_t^\prime$ from Eq.~\eqref{eq:sigmoid}, and $E_{t}, Z_{t}$ describe the current measurement or vehicle pose. The initial confidence $\ell_{o}$ is assumed to be 0, reflecting the lack of prior knowledge about the surrounding environment before the first measurement.

    \item \textit{Uncertain emerging vertex}: If $e^\prime_t$ is neither an old vertex nor trackable to an old vertex, we consider it as a potential emerging vertex corresponding to a new target in the scene. All emerging vertices are initially treated as uncertain and stored in a special set $U$ for further evaluation, aiming to reduce false alarms. Once a vertex in $U$ has been consistently associated with a new vertex more than once, it is transferred from $U$ to the polygon set $E^\prime_{t}$. The association is assumed to occur when the inter-distance is smaller than $\epsilon_2$.

    \item \textit{Missing}: This occurs when no candidate vertex is found in a sampling sector. In such cases, a virtual vertex with zero confidence is added to $E^\prime_{t}$ following the second step of the basic radar polygon formation algorithm (as outlined in Section~\ref{sec:pol_form}).
\end{itemize}
The diagram for and pseudo codes for the above are presented in Fig.~\ref{fig:poly_update} and Algorithm~\ref{alg:sim}, respectively.

\begin{algorithm}
\caption{Radar Polygon Update Algorithm}
\label{alg:sim}
\textbf{Input}: $E_{t}$ radar measurement at time $t$\\
\hspace*{2.8em} $Z_t$ vehicle pose at time $t$\\ 
\hspace*{2.8em} $L_{t-1}$ polygon vertices confidence at time $t-1$\\
\hspace*{2.8em} $E^\prime_{t-1}$ polygon vertices set at time $t-1$\\
\textbf{Output}: $E_t^\prime$ polygon vertices set at time $t$\\
\hspace*{3.2em} $L_{t}$ polygon vertices confidence at time $t$\\

\eIf{$t=1$}{        
    form basic radar polygon $E_1^\prime$, following Section~\ref{sec:pol_form}
    }{
    update $E^\prime_{t-1}$ with $Z_t$, do $E_t \gets E_t \cup E^\prime_{t-1}$ \;
    
    \For{each sampling set $S_t$ in $E_t$}{
    * find the closet candidate vertex $e^\prime_{t}$ from $S_t$ \;
    
    \If{\textbf{not} exist $e^\prime_{t}$}{
    create a virtual vertex with confidence 0\;
    
    \textbf{continue}
    }

    \If{$\left(e^\prime_{t} \textbf{ in } E^\prime_{t-1} \textbf{ and } \ell_{t-1} (e^\prime_{t}) < 0\right)$ \textbf{or} its spatial occupancy evidence $\tilde{p}$ $<p_{\text{thr}}$}{
    remove $e^\prime_{t}$ from $S_t$\;
    
    \textbf{goto} line with * and re-find a candidate vertex
    }
    
    \eIf{$e^\prime_{t}$ \textbf{in}  $E^\prime_{t-1}$}{
    $\ell_{t}(e^\prime_{t}) \gets \ell_{t}(e^\prime_{t}) - \ell_\R{pen}$
    }{
        \eIf{$e^\prime_{t}$ tracked to previous vertex $e^\prime_{t-1}$}{
        $\ell_{t}(e^\prime_{t}) \gets \ell_{t-1}(e^\prime_{t-1})+\log \frac{\tilde{p}}{1-\tilde{p}}-\ell_{o}$
        }{
        \eIf{$e_t^\prime$ trackable among $U$ and age $>$ 1}{  
        $\ell_{t}(e_t^\prime) \gets \log \frac{\tilde{p}}{1-\tilde{p}}-\ell_{o}$\;
        
        add $e^\prime_{t}$ to $E^\prime_{t}$ and add $l_{t}$ to $L_{t}$
        }{
        add $e_t^\prime$ to $U$ and update $U$
        }
        \textbf{continue}
        }
    }
    add $e^\prime_{t}$ to $E^\prime_{t}$ and add $l_{t}$ to $L_{t}$
    }
    }
\end{algorithm}




\section{Deformable Propriety of Radar Polygon \label{sec:poly_pred}}

\subsection{Deformable Polygon Prediction}
\begin{figure}
\centering
\includegraphics[width=0.48\textwidth, trim=1 2 1 1,clip]{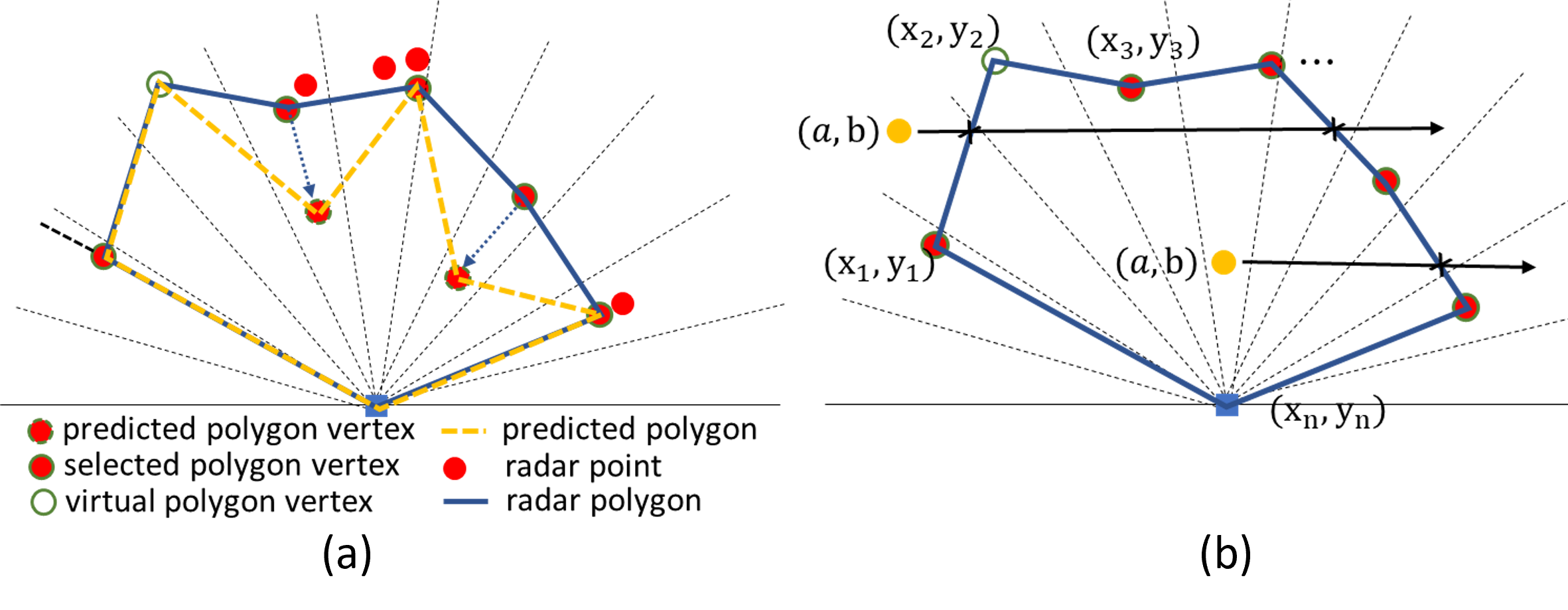}
\caption{(a) Example of a deformable radar polygon (i.e., polygon prediction); (b) Example of checking collision detection for point $(a, b)$.}
\label{fig:polygon_deform}
\end{figure}

The vertices comprising the radar polygon possess Doppler velocities, describing their instantaneous movement along the radial direction (i.e., the direction from the vertex to the radar). Under the assumption that \textit{the radial velocity of a vertex is constant within a short period and the ego-vehicle has no significant rotation}, we can estimate the vertex's future location by calculating and adding its radial movement using the current Doppler velocity (see Fig.~\ref{fig:polygon_deform}(a)). Specifically, for a vertex with location $(x_0,y_0)$ and Doppler velocity $\mathbf{v}=(v_\R{x}, v_\R{y})$, its radial movement $(\Delta x, \Delta y)$ within a duration $\Delta t$ is given by:
\begin{equation}
\label{equ:pred}
(\Delta x,\Delta y)=(v_\R{x}, v_\R{y}) \times \Delta t 
\end{equation}
\noindent where $(v_\R{x}, v_\R{y})$ is the projection of $\mathbf{v}$ along $x, y$ direction as follows:
\begin{equation}
(v_\R{x},v_\R{y}) = \frac{(x_0-x_\R{s},\ y_0-y_\R{s})}{\sqrt{(x_0-x_\R{s})^2+(y_0-y_\R{s})^2}} \times  \mathbf{v}
\end{equation}
\noindent where $(x_\R{s},y_\R{s})$ is the radar sensor location.

By adding the estimated radial movement to the current vertex, we can predict its new location $(x', y')=(x,y)+(\Delta x,\Delta y)$. The special virtual vertex, with zero Doppler velocity, remains at its original location. After predicting the new locations of all vertices in the current polygon and connecting them sequentially, we form the predicted polygon for $\Delta t$ time later. Given that the polygon shape change can be roughly and manually predicted based on Doppler velocity, we refer to this type of radar polygon as a \textit{``deformable polygon''}.

\subsection{Alternative to Pose Information for Radar Polygon Update}
The radar polygon update algorithm in Section~\ref{sec:pol_ism} relies on accurate vehicle pose information for projecting radar measurements from different time slots onto the same coordinate. Specifically, Eq.~\eqref{eq:bayes_log} illustrates the use of vehicle pose $Z_t=(x_t, y_t, \phi_t)$ in projecting the old radar polygon $E^\prime_{t-1}$ to the global coordinate of time $t$ to obtain the updated polygon. We observe that the deformable polygon prediction property—where the future shape of a radar polygon can be roughly predicted with Doppler velocity—has the potential to replace the required pose information \textit{under certain critical cases}. This substitution could be particularly useful when localization sensors (e.g., inertial measurement unit, global positioning system) are unavailable, and there is no significant lateral movement. The cross-time compensation for vehicle motion $(x_t-x_{t-1}, y_t-y_{t-1})$ and orientation change $\phi_t-\phi_{t-1}$ in Section~\ref{sec:pol_ism} can be approximately replaced by moving the vertex by $\mathbf{v}{t-1}\Delta t$, where $\mathbf{v}{t-1}$ is the corresponding Doppler velocity of the vertex, and $\Delta t$ is the time slot duration. It's important to note that Doppler velocity measures the relative moving speed along the radial direction. Even for stationary objects, non-zero Doppler velocities reflect relative movement when the car is in motion.

\section{Collision Detection for Radar Polygon \label{sec:cd}}
In this section, we address collision detection for the radar polygon-based occupancy representation. We assume the polygon has $n$ vertices $(x_1, y_1),\ (x_2, y_2),\dots, (x_n,y_n)$ sorted in ascending order of azimuth. The goal is to determine whether a point $(a,b)$, which needs to be checked, collides with the region outside the polygon. This collision detection is equivalent to determining whether the point $(a,b)$ in the plane lies inside or outside the polygon.

With that being said, we adopted the even–odd rule algorithm \cite{10.1145/368637.368653} for solving the collision detection problem. To elaborate, we draw a ray to the right of point $(a,b)$ and extend it to infinity. The number of times this ray intersects with the sides of the polygon (assuming the polygon boundary is not strictly horizontal) is then counted. As shown in Fig.~\ref{fig:polygon_deform}(b), if the ray intersects the sides an even number of times, $(a,b)$ is considered outside the polygon; if it's an odd number of times, $(a,b)$ is considered inside the polygon. A distinctive case arises when the intersection between the ray and the sides of the polygon corresponds to a vertex of the polygon. In this instance, the intersection is taken into account only if the other vertex of the side lies below the ray. The pesudo code of the collision detection algorithm for radar polygon is presented in Algorithm~\ref{alg:col_det_alg} below.

\begin{algorithm}
\caption{Collision Detection Algorithm for Radar Polygon}
\label{alg:col_det_alg}
\textbf{Input}: $(x_1, y_1),\dots, (x_n,y_n)$ polygon vertices \\
\hspace*{2.8em} $(a, b)$ any location\\
\textbf{Output}: collision indicator

\For{$i=1, 2, \dots n-1$}{
    \If{the ray starting from $(a,b)$ to its right intersects the side $(x_i, y_i)\rightarrow(x_{i+1}, y_{i+1})$}{
        \eIf{intersection is not vertex}{
            \textit{count} $+= 1$
        }{
        \If{the other vertex of the side lies below the ray}{
            \textit{count} $+= 1$
        }
        }
    }
}
\eIf{count is even}{
    \textbf{return} collision
}{
    \textbf{return} no collision
}
\end{algorithm}

It is worthy noting that by utilizing the deformable polygon property, we can predict future collision detection by forecasting the new polygon shape. According to Eq.~\eqref{equ:pred}, the future polygon shape can be roughly predicted with the input of the current polygon shape and Doppler velocity components along $x$ and $y$. Subsequently, Algorithm~\ref{alg:col_det_alg} can be applied to the predicted polygon shape for potential collision detection in the future.

\section{Implementation}
\subsection{Testbed Setup}
\begin{figure}
\centering
\includegraphics[width=0.45\textwidth, trim=1 2 0 1,clip]{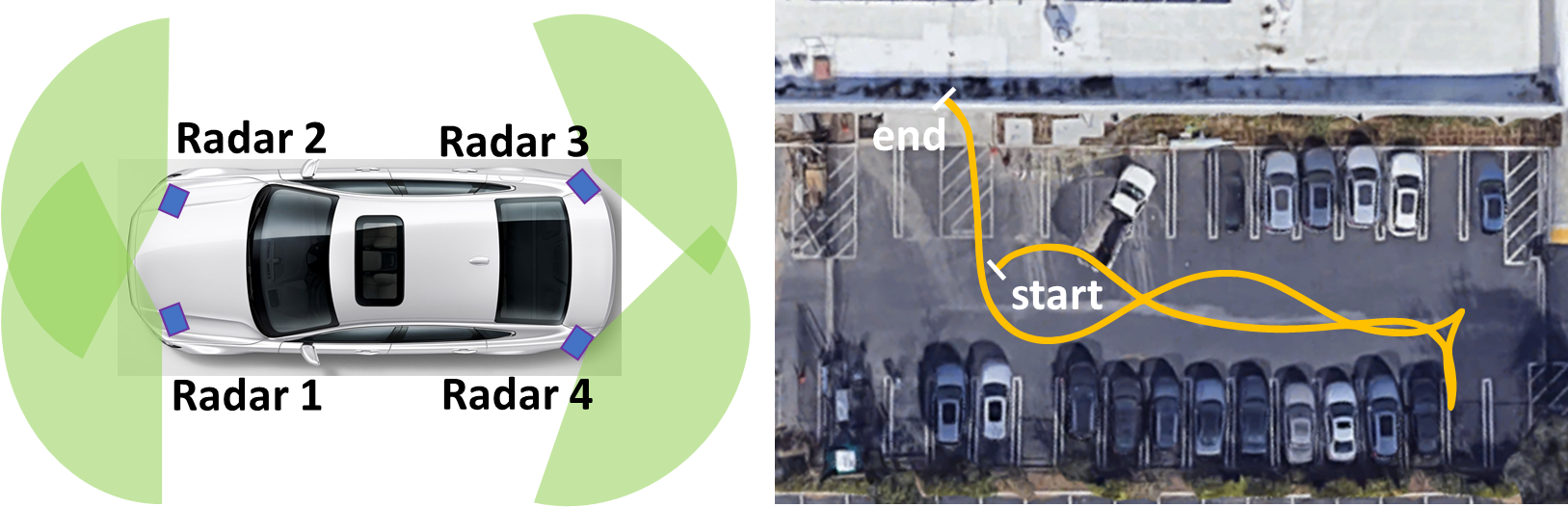}
\caption{Left: Testbed with four radars (marked as blue rectangles), with their FOVs visualized in light green. Right: BEV parking lot where the car backing-off experiment was performed along the yellow trajectory (from Google Maps).}
\label{fig:testbed}
\end{figure}

Our data collection testbed comprises four radars—two mounted on the front side and two on the rear side of the car. The radars are strategically positioned with a certain shift on boresight to collectively enhance the overall FOV, illustrated by the light green region in Fig.~\ref{fig:testbed} (left). Indeed, the radar setup depicted in Fig.~\ref{fig:testbed} does not cover the entire range, particularly the left/right sides of the ego vehicle. The two proposed polygon formation algorithms have different ways of dealing with the side area. The basic radar polygon formation algorithm leaves the sides of the vehicle blank as those areas fall outside the field of view of the radars. In contrast, the probabilistic and tracking-based radar polygon update algorithm utilizes information from previous frames, allowing for the transformation of polygon vertices to the next frame. This mechanism enables the polygon to fill or observe the sides of the ego vehicle across time as the vehicle moves.

The radar configuration includes a maximum detectable range of \SI{30}{m}, range resolution of \SI{4.3}{cm}, azimuth FOV of $\pm \ang{65}$, azimuth resolution \SI{5}{\degree}, elevation FOV of $\pm \ang{60}$, elevation resolution \SI{5}{\degree}, Doppler resolution of \SI{0.22}{m/s}, maximum unambiguous Doppler of \SI{7.1}{m/s}, and a frame rate of \SI{10}{fps}. The experimental car is also equipped with BEV cameras to facilitate synchronized data collection and ground truth labeling.

\subsection{Experiment Setup}
\label{sec:self_collect_exper}
We conducted experiments in a parking lot, simulating a classic car parking scenario that includes multiple parked vehicles and various static background elements such as walls, fences, and gates. The ego vehicle was moving at speeds of less than \SI{15}{mph} while parking in and out of spots and driving around the parking lot. The vehicle's trajectory highlighted with yellow curves in Fig.~\ref{fig:testbed} (right). The dynamic elements in the environment included pedestrians and other moving vehicles passing by as the ego vehicle was parking. These conditions were designed to mimic real-world parking scenarios, capturing both static and dynamic elements to test the robustness of our method. During the experiment, point clouds from four radars, synchronized with camera images, were collected, each with its own timestamp. Vehicle pose and trajectory, crucial for ISM, were obtained through radar odometry techniques \cite{gao2021mimosar,6907064}. Ground truth for occupancy status was generated by human labeling on BEV camera images, projected to radar coordinates via BEV-radar cross-calibration and transformation. Hyperparameters for radar polygon formation and ISM were selected through grid search: $\epsilon_1=1$, $\epsilon_2=1$, $l_\R{thr}=\SI{7.5}{m}$, $\bar{p}=12.1$, $\sigma_p=7.132$, $p_\R{thr}=0.62$, $\ell_\R{pen}=0.5$. Additional system parameters included a sampling angle of $\Delta \theta=2\degree$ and a constant false alarm rate of $p_{\R{fa}}=10^{-3}$. For computational efficiency, we formed a front-view polygon for the two front radars and a rear-view polygon for the two rear radars, calculating the union of these polygons as the final free space. All experiments were conducted on a PC equipped with a \SI{2.9}{GHz} 6-Core Intel i9 processor and \SI{32}{GB} of RAM for fair algorithmic comparisons.
\begin{figure*}
\centering
\includegraphics[width=0.98\textwidth, trim=1 3 1 1,clip]{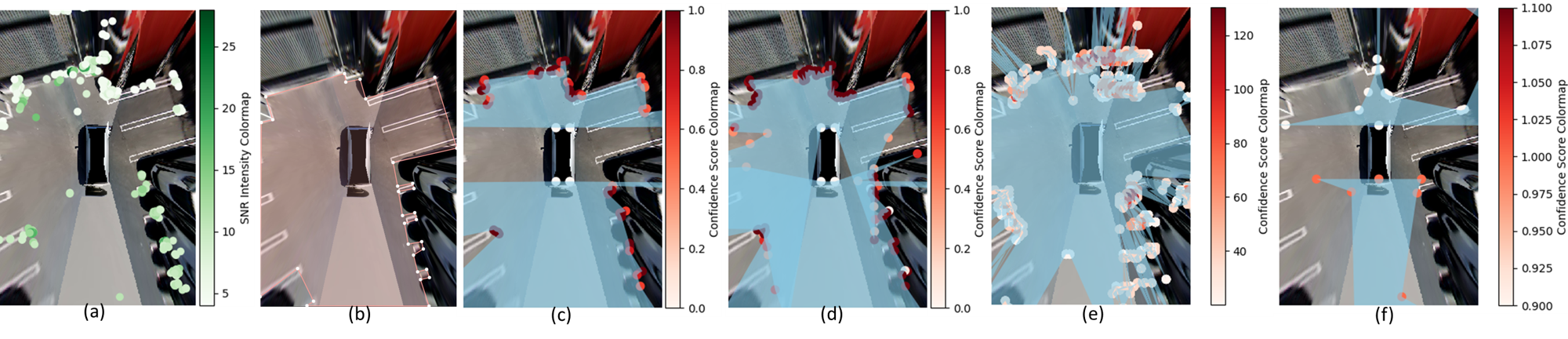}
\caption{Examples in the self-collected experiment: (a) point cloud visualization (color of points represent their SNRs), (b) free space ground truth (in red), (c) basic radar polygon result (free space in blue, polygon vertices in red points and the strangeness of red represent their confidence), (d) radar polygon update result, (e) Meerpohl \textit{et al.} \cite{MEERPOHL2019368} result, (f) Ziegler \textit{et al.} \cite{6856581} result.}
\label{fig:res_example}
\end{figure*}

\section{Evaluation and Analysis}

\subsection{Evaluation Results on Self-collected Dataset \label{sec:eva}}
\subsubsection{Baselines}
In this section, we evaluated the two proposed polygon-based algorithms on the self-collected dataset (described in Section~\ref{sec:self_collect_exper}) by measuring the accuracy and smoothness of the generated radar polygon. For simplicity of representation, we refer to the two proposed radar polygon approaches - the basic radar polygon formation algorithm, and the probabilistic and tracking-based radar polygon update algorithm - as the \textit{``basic radar polygon"} and \textit{``radar polygon update"}, respectively. 

We chose two state-of-the-art polygon-based baselines - Meerpohl \textit{et al.} \cite{MEERPOHL2019368}, and Ziegler \textit{et al.} \cite{6856581} - for comparison with the proposed algorithms. Specifically, the Meerpohl \textit{et al.} method \cite{MEERPOHL2019368} requires forming the grid map first and then raycasting on it to create the free space polygon. In our implementation, we form a grid map of $\SI{60}{m}\times\SI{60}{m}$ with $\SI{0.3}{m}$ resolution using the Werber \textit{et al.} algorithm \cite{7117922} every 30 frames. The raycasting center is chosen as the vehicle's center, the maximum distance for the raycasting line is \SI{30}{m}, the radius of the half-circle is \SI{2}{m}, the total number of sampling points is 180, and the threshold for deciding an occupied pixel is \SI{20}{dB}. The other baseline, the Ziegler \textit{et al.} algorithm \cite{6856581}, forms two convex hulls to envelop the radar point clouds on the left and right side of the vehicle separately. Since the inside of the convex hulls represents the occupancy area, we then customize the generation of the free space polygon by finding the contour of the area outside the occupancy convex hulls.

\subsubsection{Metrics}
For numerical evaluation, we employed intersection over union (IoU) as the metric. IoU is a metric used for evaluating the similarity between two shapes, such as rectangles or polygons \cite{gloudemans2023polygon}. The IoU between two polygons, e.g., A and B, is calculated by dividing the area of their intersection by the area of their union. 
Based on this, we propose two metrics for evaluation: \textit{IoU-gt} and \textit{IoU-smooth}. IoU-gt measures the IoU between the ground truth polygon and the radar polygon output from our proposed algorithms. IoU-smooth measures the IoU between the radar polygon outputs for two consecutive frames. For both IoU-gt and IoU-smooth, the values are averaged over all frames in the testing data to obtain the final results.

In addition to the IoU-gt and IoU-smooth metrics, we propose a new metric called \textit{confidence over time} (CoT) to evaluate the confidence of polygon vertices. To calculate CoT, we sum the confidence values of all vertices that match the ground truth polygon and then divide this sum by the number of vertices. This process is repeated over time to obtain the average CoT. A vertex is considered to match the ground truth polygon if the distance between it and a corresponding ground truth vertex is smaller than a specified threshold.


\subsubsection{Results}
We showcase a snapshot of recorded data, point cloud visualization, free space ground truth, and the output of the two proposed polygon algorithms and two baselines in Fig.~\ref{fig:res_example}. From it, we can observe that the basic radar polygon algorithm accurately delineates the contours of parked cars and the empty parking space between them. However, the left and right sides of the ego vehicle are left blank because these areas are outside the radars' FOV. Beyond that, the radar polygon update algorithm not only depicts the contour of the surrounding environment accurately but also fills out-of-view gaps and minimizes missing detections by utilizing temporal information. The counterpart baseline Meerpohl \textit{et al.} \cite{MEERPOHL2019368} also takes advantage of temporal information to form the free space polygon but exhibits more false alarms of vertices due to the accumulation of confidence. The Ziegler \textit{et al.} \cite{6856581} baseline, however, tends to detect smaller free space because the occupancy convex hulls usually occupy a lot of space that indeed belongs to free space. 

The numerical evaluation of the four algorithms in terms of the IoU-gt and IoU-smooth metrics is shown in Table~\ref{tab:nume_evl}. The results indicate that the basic radar polygon method achieves a 66.21\% IoU-gt and a perfect 86.45\% IoU-smooth. The adoption of the radar polygon update significantly improves IoU-gt and IoU-smooth by 7\% and 3\%, respectively. However, the two baselines perform much worse than the two proposed polygon algorithms. The Meerpohl \textit{et al.} baseline \cite{MEERPOHL2019368} has a low IoU-gt and IoU-smooth because it takes time to accumulate confidence. Thus, for early frames, the confidence of grid pixels does not reach the threshold to form radar vertices. After accumulation over a long time, the false alarm vertices also hurt its performance. The Ziegler \textit{et al.} baseline \cite{6856581} performs poorly because the generated free space polygon is much smaller than the real one. In terms of the CoT metric, the radar polygon update algorithm and the algorithm by Meerpohl \textit{et al.} \cite{MEERPOHL2019368} show higher vertex confidence compared to the other two algorithms, as they leverage data from previous frames. The Meerpohl \textit{et al.} \cite{MEERPOHL2019368} algorithm has a CoT value that is 0.04 higher, likely due to more false alarms resulting in more vertices matching the ground truth.


\begin{table}
\centering
\caption{The evaluation results on the self-collected dataset.}
\setlength\tabcolsep{3pt} 
\begin{tabular}{lccc}
\toprule  
Method &IoU-gt & IoU-smooth & CoT\\
\midrule  
basic radar polygon & 66.21\% & 86.45\%  & 0.6420\\
radar polygon update & 73.12\% & 89.68\%  & 0.7774\\
Meerpohl \textit{et al.} \cite{MEERPOHL2019368} & 40.98\% & 51.11\% & 0.8153 
\\
Ziegler \textit{et al.} \cite{6856581}  &  37.94\% & 59.50\% & 0.4852\\ 
\bottomrule 
\label{tab:nume_evl}
\end{tabular}
\end{table}




\subsection{Polygon Prediction Evaluation Results}
\begin{figure}
\centering
\includegraphics[width=0.47\textwidth, trim=1 2 1 1,clip]{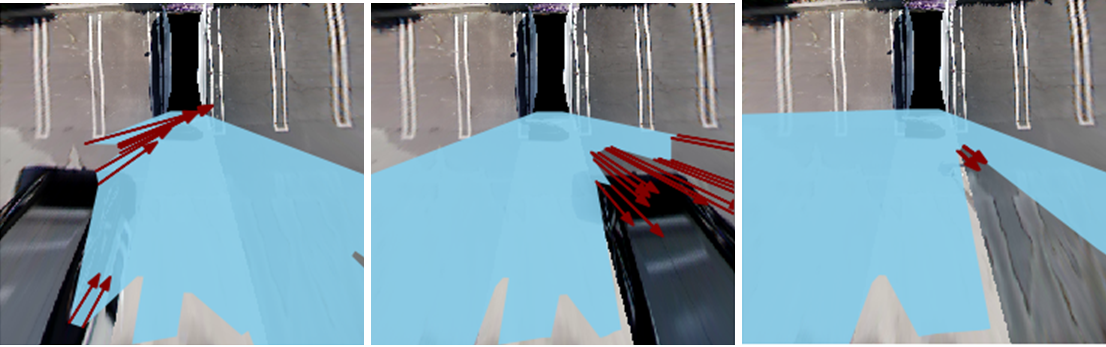}
\caption{Visualization of the Doppler velocity of the vertices: the radar polygon formed by rear radars for moving vehicles and pedestrians, and the Doppler velocity is visualized with the red arrow indicating its motion direction and speed.}
\label{fig:pedbev}

\includegraphics[width=0.45\textwidth, trim=1 2 1 1,clip]{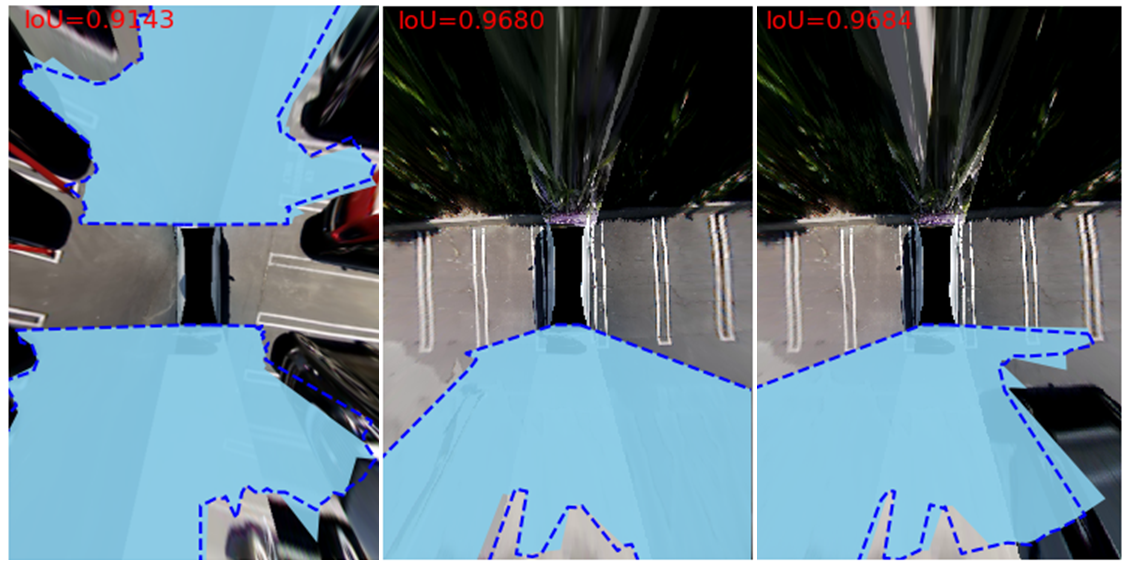}
\caption{Evaluation results of polygon prediction on the self-collected dataset. The three images represent the car backing-off, pedestrian passing-by, and vehicle passing-by experiments, respectively. The light blue region represents the measured radar polygon for current frame while the prediction from last frame is in dotted line.}
\label{fig:pred_res}
\end{figure}

To illustrate Doppler velocity on each vertex within a radar polygon, additional data was collected for two scenarios: \textit{pedestrians passing by} and \textit{vehicles passing by}. Fig.~\ref{fig:pedbev} showcases the results formed by rear radars, with red arrows indicating measured Doppler velocities for polygon vertices. The visualizations demonstrate the polygon's effective representation of the edges of passing pedestrians and moving vehicles, with the plotted Doppler velocities conveying the speed and direction of each vertex.

To quantitatively assess the performance of the polygon prediction discussed in Section~\ref{sec:poly_pred}, we conducted a one-to-one comparison in Fig.~\ref{fig:pred_res} between the current-frame radar polygon and the predicted polygon from the last frame. We calculated the IoU between them for three scenarios: car backing-off, pedestrian, and vehicle passing-by experiments. In Fig.~\ref{fig:pred_res}, current-frame radar polygons are represented by a light-blue region, while predicted polygons are depicted with a dotted line. The high IoU values above $0.9$ indicate excellent prediction performance, validating the accuracy of our deformable radar polygon approach.

It's important to note that the lack of a tangential component in Doppler speed may lead to inaccuracies for vertices with significant tangential speed, whether from self-movement or relative shifts.

\subsection{Running time and Memory Usage Analysis}
In this section, we initially assessed the execution time of the proposed radar polygon algorithms and the baselines in a PC environment. Subsequently, we conducted an analysis and measurement of the memory usage required for storing polygons and essential parameters. The results are presented in Table~\ref{tab:time_n_mem}.

According to Table~\ref{tab:time_n_mem}, the average running time of the two proposed polygon algorithms is approximately \SI{10}{ms}, significantly faster than the Meerpohl \textit{et al.} \cite{MEERPOHL2019368} algorithm, which spends a substantial amount of time on forming the radar grid before generating a polygon. The Ziegler \textit{et al.} algorithm exhibits the shortest running time among all, as its approach does not involve sophisticated processing of radar point clouds. 


The radar polygon features a lightweight storage solution with minimal memory consumption. In the following, we will analyze the storage memory usage for different algorithms. For the basic radar polygon, radar polygon update, and the Ziegler \textit{et al.} algorithms, only the vertex information (location, SNR, confidence) needs to be stored and output, resulting in a memory complexity of at most $\mathcal{O}(M)$, where $M$ is the number of sampling sectors (or vertices) in the FOV. In addition to storing vertices, the Meerpohl \textit{et al.} \cite{MEERPOHL2019368} algorithm requires storing the occupancy grid map and transferring it to the next time slot for accumulation. Thus, its memory usage is $\mathcal{O}(N^2+M)$, where $N$ is the number of cells in each dimension. Usually, the value of $N^2$ is much larger than $M$. For practical evaluation, we measured the memory usage in a PC environment for storing the vertices and grid maps, and the results are presented in Table~\ref{tab:time_n_mem}. The memory cost for the basic radar polygon, radar polygon update, and the Ziegler \textit{et al.} algorithms varies between \SI{1}{KB} and \SI{11}{KB} due to different properties of vertices being stored. However, they are much more memory-efficient than the Meerpohl \textit{et al.} \cite{MEERPOHL2019368} algorithm that requires around \SI{313}{KB} for storage due to not storing a grid map.


\begin{table}
\centering
\caption{Running time and storage memory usage for four algorithms measured on the self-collected dataset.}
\setlength\tabcolsep{3pt} 
\begin{tabular}{lcc}
\toprule  
Method &Running time & Storage memory usage\\
\midrule  
basic radar polygon & \SI{10.05}{ms} & \SI{10.93}{KB}\\
radar polygon update& \SI{9.87}{ms} &\SI{9.79}{KB} \\
Meerpohl \textit{et al.} \cite{MEERPOHL2019368} 
& \SI{741.62}{ms} & \SI{312.91}{KB}\\
Ziegler \textit{et al.} \cite{6856581} & \SI{2.128}{ms} & \SI{1.72}{KB}\\
\bottomrule 
\label{tab:time_n_mem}
\end{tabular}
\end{table}

\subsection{Impact of Sampling Angle $\Delta\theta$}
\begin{figure}
\centering
\includegraphics[width=0.48\textwidth, trim=1 1 1 1,clip]{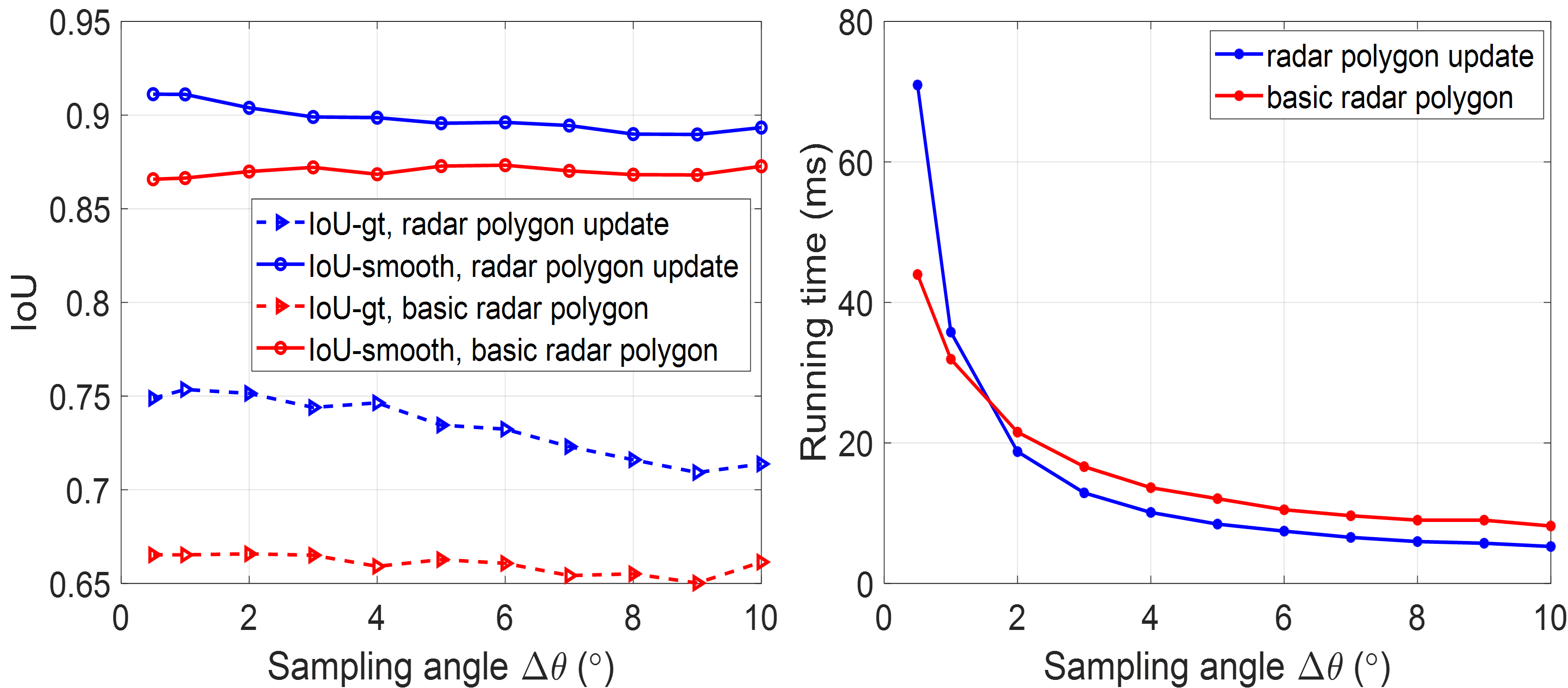}
\caption{Impact of $\Delta \theta$ on the aspect of (a) accuracy and smoothness, and (b) running time for two radar polygon algorithms.}
\label{fig:impact_samp}
\end{figure}

The sampling angle, denoted as $\Delta \theta$, significantly influences system performance and memory usage. In the experiment evaluation of the basic radar polygon and radar polygon update algorithms on the self-collected dataset, varying $\Delta \theta$ values were explored. Specifically, the sampling intervals for Fig.~\ref{fig:impact_samp} were set at 0.5, 1, 2, 3, 4, 5, 6, 7, 8, 9, and 10 degrees. As depicted in Fig.~\ref{fig:impact_samp}, a smaller $\Delta \theta$ yields more complete and accurate radar polygons, leading to improved IoU-gt and IoU-smooth for both methods. While accuracy and smoothness are not dramatically compromised with $\Delta \theta$ ranging from $0.5\degree$ to $10\degree$, diminishing the sampling angle introduces computational complexity and memory challenges due to an increased number of vertices for calculation, storage, and updating. Consequently, finding an optimal trade-off for the $\Delta \theta$ value becomes crucial to align with specific application requirements.

\begin{figure*}
\centering
\includegraphics[width=0.95\textwidth, trim=0 3 1 0,clip]{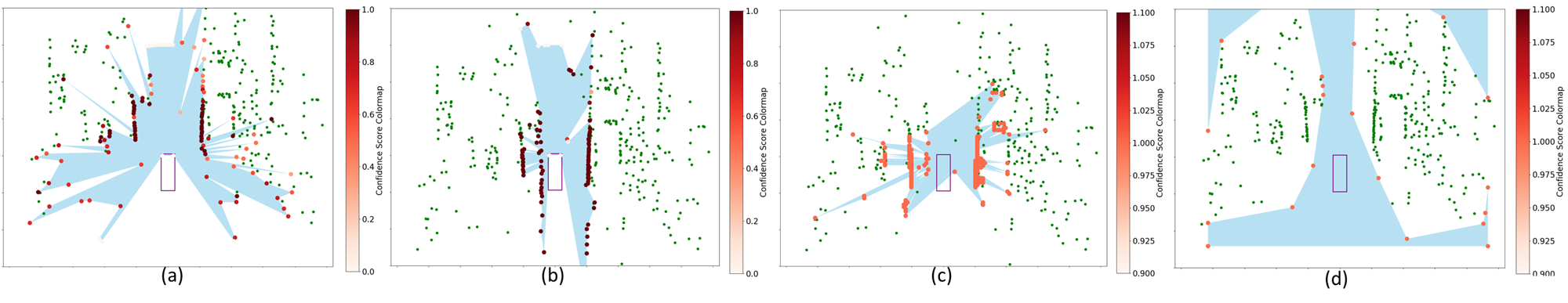}
\caption{Examples in RadarScenes dataset: (a) result of basic radar polygon (free space is in blue), (b) result of radar polygon update algorithms, (c) result of Meerpohl \textit{et al.} \cite{MEERPOHL2019368}, (d) result of Ziegler \textit{et al.} \cite{6856581}. Among all figure, the green points are input radar point clouds and the red dots are the radar polygon vertices.}
\label{fig:res_example_radarscene}
\end{figure*}

\subsection{Evaluation Results on RadarScenes Dataset}
To ensure the robustness of our proposed radar polygon algorithms, we conducted evaluations not only for the radar sensor combinations shown in Fig.\ref{fig:testbed}, representing our self-collected dataset but also on the publicly available RadarScenes dataset \cite{radar_scenes_dataset}, focusing on city-road driving scenarios. The RadarScenes dataset features four front-view radars with diverse setups and locations, distinct from our testbed. Specifically, a clip comprising 916 frames was selected from the whole dataset for analysis. Similar to the previous numerical evaluations, the IoU-gt and IoU-smooth for the two proposed radar polygon algorithms and two baselines were computed and are presented in Table.~\ref{tab:nume_evl_radarscene}. The basic radar polygon method achieved a 23.66\% IoU-gt and a 68.91\% IoU-smooth, while the radar polygon update algorithm demonstrated a significant improvement with metrics reaching 74.44\% and 86.38\%, respectively. This substantial gain is attributed to the capability of the radar polygon update algorithm to fill the out-of-view gap with old or trackable vertices and reduce false alarms from uncertain vertices. The two baselines both achieve slightly worse IoU-gt compared to the basic radar polygon algorithm but slightly better IoU-smooth. In terms of the CoT metric, the radar polygon update algorithm achieves the highest value of 0.8507, indicating the best vertex confidence. The algorithm by Meerpohl \textit{et al.} is slightly lower, while the other two algorithms have significantly lower values due to their single-frame operations.

\begin{figure}
\centering
\includegraphics[width=0.46\textwidth, trim=1 2 1 0,clip]{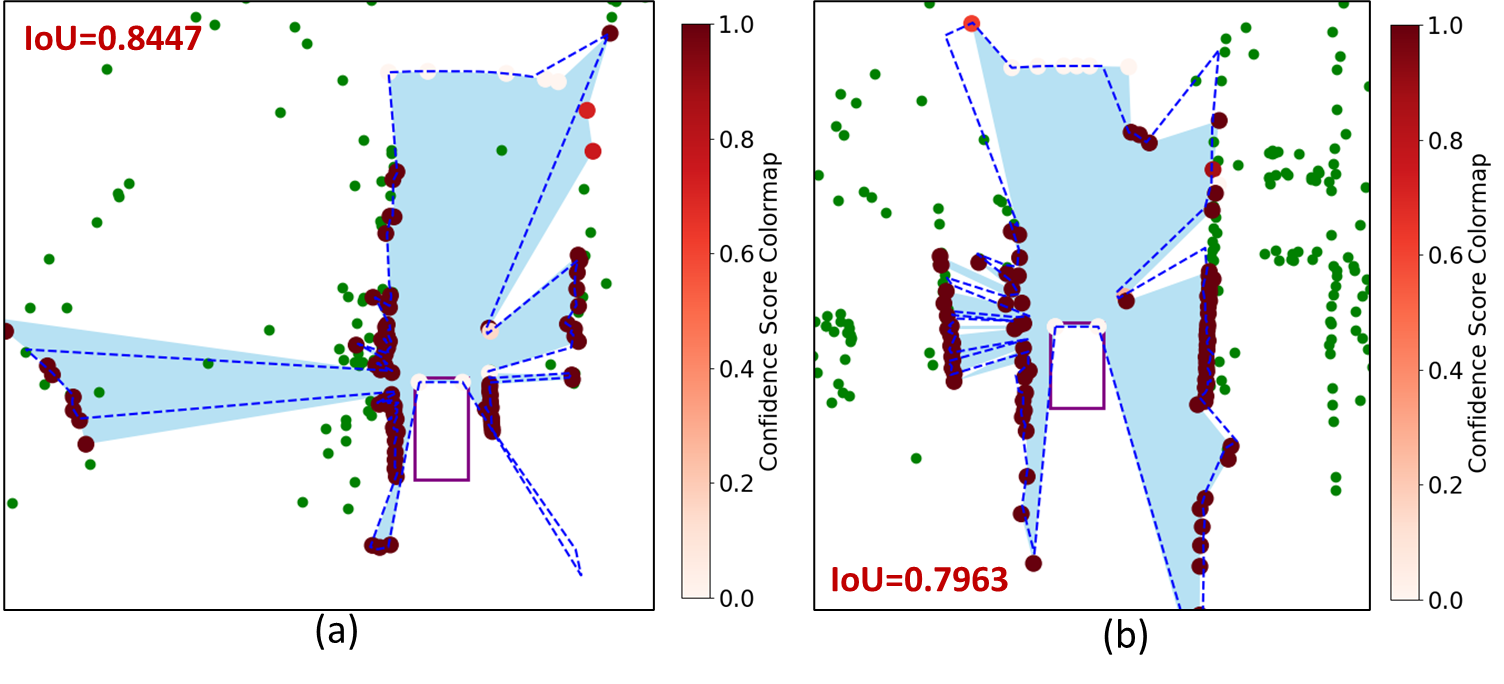}
\caption{Evaluation results of polygon prediction on the RadarScenes dataset. The light blue region is the measured radar polygon while the dotted line forms the predicted radar polygon from the last frame.}
\label{fig:pred_res_radarscene}
\end{figure}

A snapshot of recorded data along with the radar polygon outputs for the four algorithms is visualized in Fig.~\ref{fig:res_example_radarscene}, where we can observe that the polygon generated by the radar polygon update algorithm depicts the surrounding area most accurately. The evaluation results of the polygon prediction are depicted in Fig.~\ref{fig:pred_res_radarscene}. Even in the complex city-road driving scenario, the deformable polygon demonstrates robust short-term prediction performance, exhibiting a high overlap with the current-frame radar polygon. The IoU values for two testing cases are $84.47\%$ and $79.63\%$, respectively. However, a notable issue is observed, as the prediction does not effectively handle new points or objects that appear, resulting in a degradation of IoU performance due to the lack of information from previous frames.

\begin{table}
\centering
\caption{The evaluation results on the RadarScenes dataset.}
\setlength\tabcolsep{3pt} 
\begin{tabular}{lccc}
\toprule  
Method &IoU-gt & IoU-smooth  & CoT\\
\midrule  
basic radar polygon & 23.66\% & 68.91\% & 0.4402\\
radar polygon update & 74.44\% & 86.38\%  & 0.8507\\
Meerpohl \textit{et al.} \cite{MEERPOHL2019368} & 22.91\% & 70.55\% & 0.7871\\
Ziegler \textit{et al.} \cite{6856581} & 18.98\% & 73.23\% & 0.2633\\
\bottomrule 
\label{tab:nume_evl_radarscene}
\end{tabular}
\end{table}

\subsection{Collision Detection for Route Planning}
\begin{figure}
\centering
\includegraphics[width=0.36\textwidth]{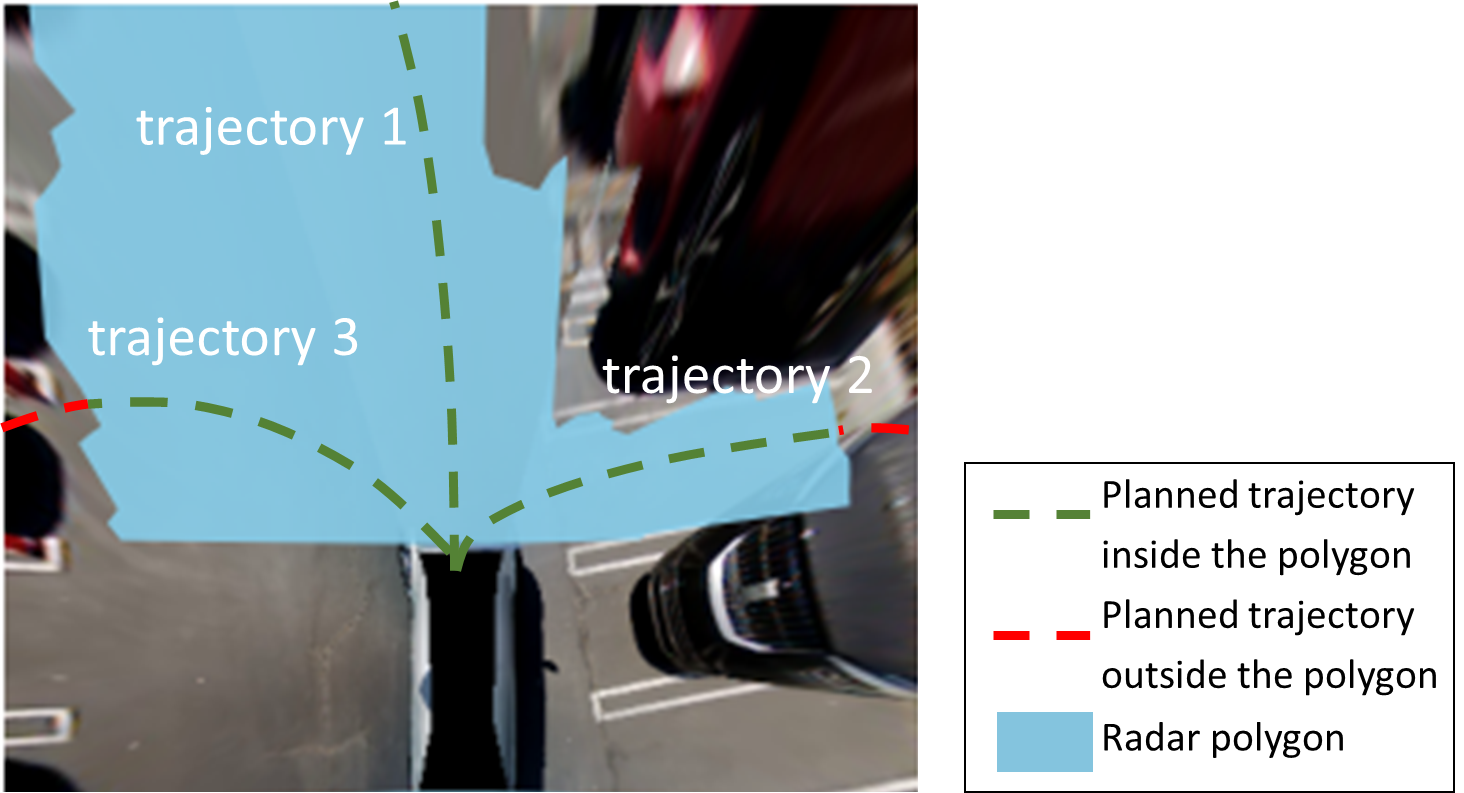}
\caption{Collision detection results for vehicle route planning.}
\label{fig:coll_res}
\end{figure}

The collision detection algorithm, as discussed in Section~\ref{sec:cd}, was employed to process a vehicle route planning scenario, assessing whether planned trajectories fall within the radar polygon. We adopted a radar polygon representing the free space for driving generated by our algorithms on the self-collected dataset, depicted in blue in Fig.~\ref{fig:coll_res}. Subsequently, three trajectories were randomly generated, and each trajectory was sampled to acquire a set of locations. These locations served as input for the collision detection algorithm, determining whether each point was inside or outside the polygon. The collision detection results for the planned trajectories are depicted in Fig.~\ref{fig:coll_res} as red or green dotted lines, with red indicating collision in this path and green representing no collision. After a thorough evaluation, we established a $100\%$ collision detection accuracy. This remarkable result underscores the effectiveness of the collision detection algorithm in ensuring the safety of planned vehicle trajectories.


\section{Conclusion}
In this paper, we introduced a radar polygon formation algorithm that utilizes radar point cloud data as input. Additionally, we developed a collision detection method tailored to this new polygon representation. Both the radar polygon and the collision detection model were validated through extensive experiments using both self-collected and public datasets. Our results show that the IoU-gt and IoU-smooth metrics increased by 35\% and around 40\%, respectively, compared to baseline methods on the self-collected dataset, and by 52\% and 13\%, respectively, on the RadarScenes dataset, demonstrating superior accuracy and robustness. Furthermore, the deformable polygon concept was validated by the high IoU correlation between the predicted and generated radar polygons. Memory usage analysis also indicated significantly reduced memory requirements compared to baselines. For future work, we plan to explore approaches to adapt the radar polygon for long-range scenarios and further enhance its overall robustness.

\bibliographystyle{IEEEtran}
\bibliography{bibtex}

\begin{thebibliography}{10}
\providecommand{\url}[1]{#1}
\csname url@samestyle\endcsname
\providecommand{\newblock}{\relax}
\providecommand{\bibinfo}[2]{#2}
\providecommand{\BIBentrySTDinterwordspacing}{\spaceskip=0pt\relax}
\providecommand{\BIBentryALTinterwordstretchfactor}{4}
\providecommand{\BIBentryALTinterwordspacing}{\spaceskip=\fontdimen2\font plus
\BIBentryALTinterwordstretchfactor\fontdimen3\font minus \fontdimen4\font\relax}
\providecommand{\BIBforeignlanguage}[2]{{%
\expandafter\ifx\csname l@#1\endcsname\relax
\typeout{** WARNING: IEEEtran.bst: No hyphenation pattern has been}%
\typeout{** loaded for the language `#1'. Using the pattern for}%
\typeout{** the default language instead.}%
\else
\language=\csname l@#1\endcsname
\fi
#2}}
\providecommand{\BIBdecl}{\relax}
\BIBdecl

\bibitem{srr}
\BIBentryALTinterwordspacing
AVNET, \emph{Automotive radar systems: the design engineer’s guide}.\hskip 1em plus 0.5em minus 0.4em\relax AVNET, 2024. [Online]. Available: \url{https://my.avnet.com/abacus/solutions/markets/automotive-and-transportation/automotive/comfort-infotainment-and-safety/automotive-radar/}
\BIBentrySTDinterwordspacing

\bibitem{gao2019experiments}
X.~Gao, G.~{Xing}, S.~{Roy}, and H.~{Liu}, ``Experiments with mmwave automotive radar test-bed,'' in \emph{2019 53rd Asilomar Conference on Signals, Systems, and Computers}, 2019, pp. 1--6.

\bibitem{gao2021mimosar}
X.~Gao, S.~Roy, and G.~Xing, ``Mimo-sar: A hierarchical high-resolution imaging algorithm for mmwave fmcw radar in autonomous driving,'' \emph{IEEE Transactions on Vehicular Technology}, vol.~70, no.~8, pp. 7322--7334, 2021.

\bibitem{ti_uss}
\BIBentryALTinterwordspacing
A.~W. Mubina~Toa, \emph{Ultrasonic Sensing Basics}.\hskip 1em plus 0.5em minus 0.4em\relax Texas Instrument, 2019, no. SLAA907C. [Online]. Available: \url{https://www.ti.com/lit/an/slaa907c/slaa907c.pdf}
\BIBentrySTDinterwordspacing

\bibitem{ramp}
X.~{Gao}, G.~{Xing}, S.~{Roy}, and H.~{Liu}, ``Ramp-cnn: A novel neural network for enhanced automotive radar object recognition,'' \emph{IEEE Sensors Journal}, vol.~21, no.~4, pp. 5119--5132, 2021.

\bibitem{gao2021perception}
X.~Gao, S.~Roy, G.~Xing, and S.~Jin, ``Perception through 2d-mimo fmcw automotive radar under adverse weather,'' in \emph{2021 IEEE International Conference on Autonomous Systems (ICAS)}, 2021, pp. 1--5.

\bibitem{8443484}
R.~Prophet, H.~Stark, M.~Hoffmann, C.~Sturm, and M.~Vossiek, ``Adaptions for automotive radar based occupancy gridmaps,'' in \emph{2018 IEEE MTT-S International Conference on Microwaves for Intelligent Mobility (ICMIM)}, 2018, pp. 1--4.

\bibitem{derive_spat_occup}
P.~Berthold, M.~Michaelis, T.~Luettel, D.~Meissner, and H.-J. Wuensche, ``Deriving spatial occupancy evidence from radar detection data,'' in \emph{2020 IEEE Intelligent Vehicles Symposium (IV)}, 2020, pp. 831--836.

\bibitem{7535495}
J.~Degerman, T.~Pernstål, and K.~Alenljung, ``3d occupancy grid mapping using statistical radar models,'' in \emph{2016 IEEE Intelligent Vehicles Symposium (IV)}, 2016, pp. 902--908.

\bibitem{9299052}
F.~E. Nowruzi, D.~Kolhatkar, P.~Kapoor, F.~Al~Hassanat, E.~J. Heravi, R.~Laganiere, J.~Rebut, and W.~Malik, ``Deep open space segmentation using automotive radar,'' in \emph{2020 IEEE MTT-S International Conference on Microwaves for Intelligent Mobility (ICMIM)}, 2020, pp. 1--4.

\bibitem{8916897}
N.~Engelhardt, R.~Pérez, and Q.~Rao, ``Occupancy grids generation using deep radar network for autonomous driving,'' in \emph{2019 IEEE Intelligent Transportation Systems Conference (ITSC)}, 2019, pp. 2866--2871.

\bibitem{9022091}
L.~Sless, B.~E. Shlomo, G.~Cohen, and S.~Oron, ``Road scene understanding by occupancy grid learning from sparse radar clusters using semantic segmentation,'' in \emph{2019 IEEE/CVF International Conference on Computer Vision Workshop (ICCVW)}, 2019, pp. 867--875.

\bibitem{7117922}
K.~Werber, M.~Rapp, J.~Klappstein, M.~Hahn, J.~Dickmann, K.~Dietmayer, and C.~Waldschmidt, ``Automotive radar gridmap representations,'' in \emph{2015 IEEE MTT-S International Conference on Microwaves for Intelligent Mobility (ICMIM)}, 2015, pp. 1--4.

\bibitem{9294626}
C.~Diehl, E.~Feicho, A.~Schwambach, T.~Dammeier, E.~Mares, and T.~Bertram, ``Radar-based dynamic occupancy grid mapping and object detection,'' in \emph{2020 IEEE 23rd International Conference on Intelligent Transportation Systems (ITSC)}, 2020, pp. 1--6.

\bibitem{Li2018HighRR}
M.~Li, Z.~Feng, M.~Stolz, M.~Kunert, R.~Henze, and F.~K{\"u}ç{\"u}kay, ``High resolution radar-based occupancy grid mapping and free space detection,'' in \emph{VEHITS}, 2018.

\bibitem{5548091}
F.~Homm, N.~Kaempchen, J.~Ota, and D.~Burschka, ``Efficient occupancy grid computation on the gpu with lidar and radar for road boundary detection,'' in \emph{2010 IEEE Intelligent Vehicles Symposium}, 2010, pp. 1006--1013.

\bibitem{8813839}
M.~Slutsky and D.~Dobkin, ``Fast implementation of volumetric occupancy grids,'' in \emph{2019 IEEE Intelligent Vehicles Symposium (IV)}, 2019, pp. 750--755.

\bibitem{gao2022learning}
X.~Gao, H.~Liu, S.~Roy, G.~Xing, A.~Alansari, and Y.~Luo, ``Learning to detect open carry and concealed object with 77 ghz radar,'' \emph{IEEE journal of selected topics in signal processing}, vol.~16, no.~4, pp. 791--803, 2022.

\bibitem{Sless_2019_ICCV}
L.~Sless, B.~El~Shlomo, G.~Cohen, and S.~Oron, ``Road scene understanding by occupancy grid learning from sparse radar clusters using semantic segmentation,'' in \emph{Proceedings of the IEEE/CVF International Conference on Computer Vision (ICCV) Workshops}, Oct 2019.

\bibitem{xie2022deepvs}
Z.~Xie, H.~Wang, S.~Han, E.~Schoenfeld, and F.~Ye, ``Deepvs: A deep learning approach for rf-based vital signs sensing,'' in \emph{Proceedings of the 13th ACM international conference on bioinformatics, computational biology and health informatics}, 2022, pp. 1--5.

\bibitem{10273590}
Y.~Jin, M.~Hoffmann, A.~Deligiannis, J.-C. Fuentes-Michel, and M.~Vossiek, ``Semantic segmentation-based occupancy grid map learning with automotive radar raw data,'' \emph{IEEE Transactions on Intelligent Vehicles}, pp. 1--16, 2023.

\bibitem{9341308}
D.~Bauer, L.~Kuhnert, and L.~Eckstein, ``Deep inverse sensor models as priors for evidential occupancy mapping,'' in \emph{2020 IEEE/RSJ International Conference on Intelligent Robots and Systems (IROS)}, 2020, pp. 6032--3067.

\bibitem{xie2021vitalhub}
Z.~Xie, B.~Zhou, X.~Cheng, E.~Schoenfeld, and F.~Ye, ``Vitalhub: Robust, non-touch multi-user vital signs monitoring using depth camera-aided uwb,'' in \emph{2021 IEEE 9th International Conference on Healthcare Informatics (ICHI)}.\hskip 1em plus 0.5em minus 0.4em\relax IEEE, 2021, pp. 320--329.

\bibitem{8793263}
R.~Weston, S.~Cen, P.~Newman, and I.~Posner, ``Probably unknown: Deep inverse sensor modelling radar,'' in \emph{2019 International Conference on Robotics and Automation (ICRA)}, 2019, pp. 5446--54s52.

\bibitem{xie2022passive}
Z.~Xie, B.~Zhou, X.~Cheng, E.~Schoenfeld, and F.~Ye, ``Passive and context-aware in-home vital signs monitoring using co-located uwb-depth sensor fusion,'' \emph{ACM transactions on computing for healthcare}, vol.~3, no.~4, pp. 1--31, 2022.

\bibitem{10.1145/368637.368653}
\BIBentryALTinterwordspacing
M.~Shimrat, ``Algorithm 112: Position of point relative to polygon,'' \emph{Commun. ACM}, vol.~5, no.~8, p. 434, Aug. 1962. [Online]. Available: \url{https://doi.org/10.1145/368637.368653}
\BIBentrySTDinterwordspacing

\bibitem{gao2023static}
X.~Gao, S.~Roy, and L.~Zhang, ``Static background removal in vehicular radar: Filtering in azimuth-elevation-doppler domain,'' \emph{arXiv preprint arXiv:2307.01444}, 2023.

\bibitem{8793503}
F.~Ruetz, E.~Hernández, M.~Pfeiffer, H.~Oleynikova, M.~Cox, T.~Lowe, and P.~Borges, ``Ovpc mesh: 3d free-space representation for local ground vehicle navigation,'' in \emph{2019 International Conference on Robotics and Automation (ICRA)}, 2019, pp. 8648--8654.

\bibitem{9740418}
M.~I. Hussain, S.~Azam, M.~A. Rafique, A.~M. Sheri, and M.~Jeon, ``Drivable region estimation for self-driving vehicles using radar,'' \emph{IEEE Transactions on Vehicular Technology}, vol.~71, no.~6, pp. 5971--5982, 2022.

\bibitem{MEERPOHL2019368}
C.~Meerpohl, M.~Rick, and C.~Büskens, ``Free-space polygon creation based on occupancy grid maps for trajectory optimization methods,'' \emph{IFAC-PapersOnLine}, vol.~52, no.~8, pp. 368--374, 2019.

\bibitem{6856581}
J.~Ziegler, P.~Bender, T.~Dang, and C.~Stiller, ``Trajectory planning for bertha — a local, continuous method,'' in \emph{2014 IEEE Intelligent Vehicles Symposium Proceedings}, 2014, pp. 450--457.

\bibitem{kuan1985natural}
D.~Kuan, J.~Zamiska, and R.~Brooks, ``Natural decomposition of free space for path planning,'' in \emph{Proceedings. 1985 IEEE International Conference on Robotics and Automation}, vol.~2.\hskip 1em plus 0.5em minus 0.4em\relax IEEE, 1985, pp. 168--173.

\bibitem{6338636}
M.~Schreier and V.~Willert, ``Robust free space detection in occupancy grid maps by methods of image analysis and dynamic b-spline contour tracking,'' in \emph{2012 15th International IEEE Conference on Intelligent Transportation Systems}, 2012, pp. 514--521.

\bibitem{6856568}
R.~Dubé, M.~Hahn, M.~Schütz, J.~Dickmann, and D.~Gingras, ``Detection of parked vehicles from a radar based occupancy grid,'' in \emph{2014 IEEE Intelligent Vehicles Symposium Proceedings}, 2014, pp. 1415--1420.

\bibitem{7918864}
R.~Prophet, M.~Hoffmann, M.~Vossiek, G.~Li, and C.~Sturm, ``Parking space detection from a radar based target list,'' in \emph{2017 IEEE MTT-S International Conference on Microwaves for Intelligent Mobility (ICMIM)}, 2017, pp. 91--94.

\bibitem{xie2021signal}
Z.~Xie, B.~Zhou, and F.~Ye, ``Signal quality detection towards practical non-touch vital sign monitoring,'' in \emph{Proceedings of the 12th ACM Conference on Bioinformatics, Computational Biology, and Health Informatics}, 2021, pp. 1--9.

\bibitem{xie2021fusing}
Z.~Xie, B.~Zhou, X.~Cheng, E.~Schoenfeld, and F.~Ye, ``Fusing uwb and depth sensors for passive and context-aware vital signs monitoring,'' in \emph{2021 IEEE/ACM Conference on Connected Health: Applications, Systems and Engineering Technologies (CHASE)}.\hskip 1em plus 0.5em minus 0.4em\relax IEEE, 2021, pp. 119--120.

\bibitem{6907064}
D.~Kellner, M.~Barjenbruch, J.~Klappstein, J.~Dickmann, and K.~Dietmayer, ``Instantaneous ego-motion estimation using multiple doppler radars,'' in \emph{2014 IEEE International Conference on Robotics and Automation (ICRA)}, 2014, pp. 1592--1597.

\bibitem{gloudemans2023polygon}
D.~Gloudemans, X.~Lu, S.~Xia, and D.~B. Work, ``Polygon intersection-over-union loss for viewpoint-agnostic monocular 3d vehicle detection,'' \emph{arXiv preprint arXiv:2309.07104}, 2023.

\bibitem{radar_scenes_dataset}
\BIBentryALTinterwordspacing
O.~Schumann, M.~Hahn, N.~Scheiner, F.~Weishaupt, J.~Tilly, J.~Dickmann, and C.~Wöhler, ``{RadarScenes: A Real-World Radar Point Cloud Data Set for Automotive Applications},'' Mar. 2021. [Online]. Available: \url{https://doi.org/10.5281/zenodo.4559821}
\BIBentrySTDinterwordspacing

\end{thebibliography}

\end{document}